# Population-Guided Large Margin Classifier for High-Dimension Low -Sample-Size Problems

Qingbo Yin[1,2], Ehsan Adeli[3], Liran Shen[4], Dinggang Shen[2,5, *]

[1] College of Information Science and Technology, Dalian Maritime University, Dalian, 116023, China

[2] Department of Radiology and BRIC, University of North Carolina, Chapel Hill, NC 27599, USA

[3] Departments of Psychiatry & Behavioral Sciences and Computer Science, Stanford University, Stanford, CA 94305-5723, USA

[4] College of Marine Electrical Engineering, Dalian Maritime University, Dalian, 116023, China

[5] Department of Brain and Cognitive Engineering, Korea University, Seoul 02841, Republic of Korea

*Correspondence and requests for materials should be addressed to Dinggang Shen (Email: dgshen@med.unc.edu)

**Abstract:**

Various applications in different fields, such as gene expression analysis or computer vision, suffer from data sets with high-dimensional low-sample-size (HDLSS), which has posed significant challenges for standard statistical and modern machine learning methods. In this paper, we propose a novel linear binary classifier, denoted by population-guided large margin classifier (PGLMC), which is applicable to any sorts of data, including HDLSS. PGLMC is conceived with a projecting direction *w* given by the comprehensive consideration of local structural information of the hyperplane and the statistics of the training samples. Our proposed model has several advantages compared to those widely used approaches. First, it is not sensitive to the intercept term *b*. Second, it operates well with imbalanced data. Third, it is relatively simple to be implemented based on Quadratic Programming. Fourth, it is robust to the model specification for various real applications. The theoretical properties of PGLMC are proven. We conduct a series of evaluations on two simulated and six real-world benchmark data sets, including DNA classification, digit recognition, medical image analysis, and face recognition. PGLMC outperforms the state-of-the-art classification methods in most cases, or at least obtains comparable results.

# 1. Introduction

Classification, as a supervised-learning tool, has always been an integral part of various applications of statistics and machine learning to real-world problems, such as in computer vision, medical image analysis, knowledge discovery, and pattern recognition. Despite its extensive research and widespread applications, when confronted with complex problems (such as dealing with genetic microarrays[1-3], medical images[4, 5], and chemometrics [6, 7]), there are still some unresolved challenges. These challenges include: 1) High dimensionality of the data while having too few samples for training (for short, high-dimensional low-sample-size, HDLSS), where the sample size, *n*, is much smaller than the feature dimension *d* ($n << d$); 2) Imbalanced data (i.e., skewed distribution of samples across classes); and 3) Presence of noisy data in the samples. These factors often degrade the performance of the classifiers.

A typical example of HDLSS is on gene expression that a single measurement yields simultaneous expression levels for thousands to tens of thousands [8]. As acquiring samples for such applications is very costly, the number of samples in such data sets is much fewer than the dimensionality of each sample (i.e., $n << d$). At the same time, it is inevitable in HDLSS to acquire the imbalanced data across the classes.

Although there is a great advancement on classification methods in different contexts, the HDLSS data has posed significant challenges to standard statistical methods and have rendered many existing classification techniques impractical [9]. The related studies on HDLSS problem can be categorized into two main groups. The first group involves the methods requiring specific preprocessing steps such as regularity or dimensionality reduction in the feature space. This group of methods includes the majority of modern classifiers, i.e., Mean Difference (MD) [10], Logistic Regression (LR) [11], Naïve Bayes (NB) [10], Linear Discriminant Analysis (LDA) [12], ensemble methods [13], Neural Network (NN), and deep learning [14]. Although sometimes they work well in specific scenarios, these classifiers often suffer from serious drawbacks of having biased discriminant scores which were derived from the assumption of feature independence and unstable estimation of high-dimensional covariance matrices, and even being infeasible when the dimension of data and the sample size are very large [15, 16]. The second group only contains a few of methods, which can be directly applied for classification on HDLSS data sets without consideration of any dimensionality reduction. In this group of methods, two well-known methods with geometry representation, i.e., Support Vector Machine (SVM) [17] and the distance-weighted methods [18-21] (i.e., Distance-Weighted Discrimination (DWD), weighted Distance-Weighted

Discrimination (wDWD), and Distance-Weighted Support Vector Machine (DWSVM)) are the state-of-the-art approaches. In this paper, we only focus on the methods without the need to conduct any dimensionality reduction, as these methods are the core approaches to deal with HDLSS and can be integrated with any other pre- or post-processing steps.

In the HDLSS data setting, a so-called “data-piling” phenomenon is often observed for SVM and some other classifiers [19, 22]. Data-piling refers to a large portion of the data lying on two hyperplanes or concentrate on two points after the data are projected to the direction vector given by the linear classifier [22]. The phenomenon of data-piling indicates severe overfitting in the HDLSS data setting. On the other hand, although DWD is a recently developed classifier to improve SVM and overcome the data-piling issue in the HDLSS data setting, it is sensitive to the imbalanced sample sizes across the two classes. Further improvement methods for DWD, including wDWD and DWSVM, conquer this shortcoming. However, these methods derived from DWD are more computationally expensive due to the second-order cone programming (SOCP) compared to the quadratic programming for SVM [23-25].

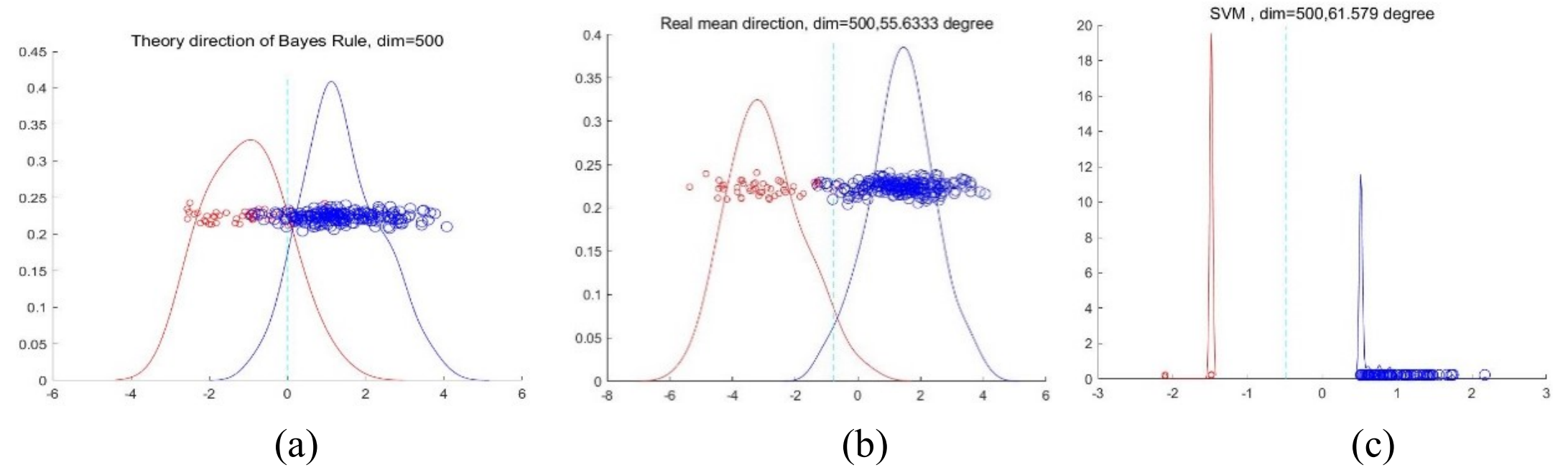

(a) (b) (c)

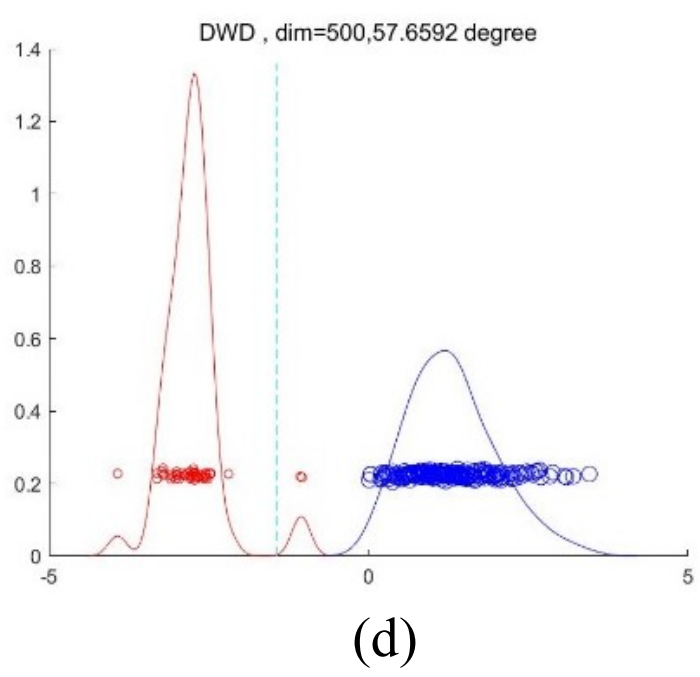

(d)

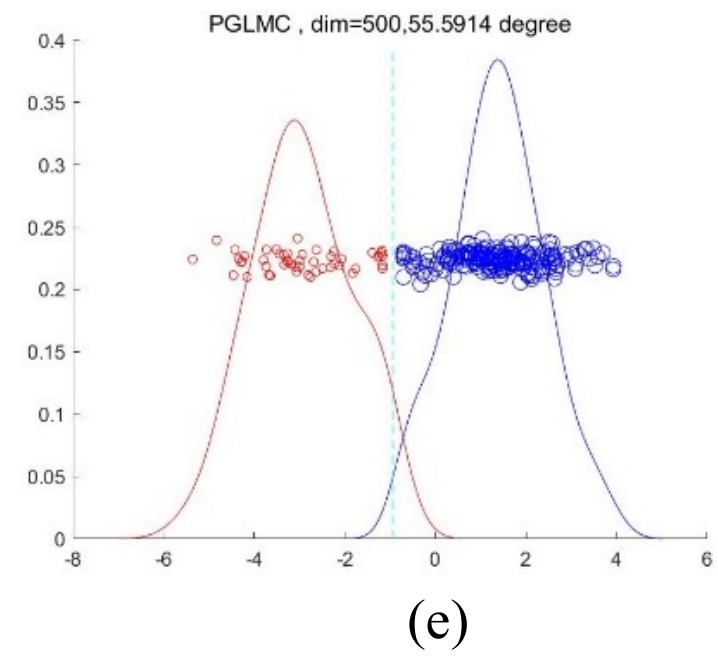

(e)

Figure 1. Plots of projections, discriminant planes and the density estimate for different methods. (a) The true theory direction of Bayes rule. (b) Real mean difference direction for two classes. (c) SVM direction. (d) DWD direction. (e) The proposed PGLMC direction.

Both of data-piling and imbalanced data adversely affect the classification model. A toy example is depicted for the issue of data-piling and imbalanced data in Fig.1. This figure plots the samples in two-dimensional spaces, and projects these samples to three different discriminant directions. All data from two classes are generated from a multivariate normal distribution $N_d(\pm \mu,\Sigma)$, where $\mu \equiv c\mathbf{1}_d$, $2c\|\mathbf{1}_d\|_2 = 2.7$ and $\Sigma \equiv I_d$, the dimension $d = 500$, $\mathbf{1}_d$ is a $d$-dimensional vector of all ones, and $I_d$ is a $d \times d$ identity matrix. The Bayes rule in this example has the direction $w_{Bayes} = \Sigma^{-1}(2\mu) = \mathbf{2.7}*\mathbf{1}_d/\|\mathbf{1}_d\|$, and the Bayes intercept is $b_{Bayes} = 0$. The sample size of the positive class (blue points in the figure) is $n_+ = 200$, and the sample size of the negative class (red points) sample size is $n_- = 50$. We adopt the sample size ratio $m \geq 1$ as the imbalanced factor between the majority class and minority class. Here, $m = 4$. Figure 1(a) illustrates the theory direction of Bayes rule derived from the generating distribution, and the density estimate for the projection of all data vectors. Here, the Bayes direction [6], which serves as the benchmark to be compared with, is only theory direction, which is defined in Subsection 4.1. Whereas, Figure 1(b) shows the real mean difference direction of two classes, and shows the deviation which is measured by the angle between real mean difference direction and Bayes rule direction. This real mean difference direction is derived from all sample vectors, using the sample mean of two classes, instead of $\mu$ in the same equation with Bayes direction. The intercept for this direction is the median of two means from two class samples. In real applications, since the theory distribution of samples cannot be known, the real mean difference direction of two classes is a valuable benchmark, instead of the theoretical direction. Figure 1(c) shows the SVM direction, in which all samples can be clearly separated into two classes. It demonstrates an interesting phenomenon, data-piling, that a large portion of the data pile up and lie on two hyperplanes parallel to the SVM classification boundary. This is an obvious indicator for overfitting. The SVM model does not provide any clues on the real data distribution. Figure 1(d) shows that DWD avoid the data-piling issue and the data projection maintain the Gaussian pattern, which implies that there is some potential to interpret the data and generalize this model to new data in terms of DWD direction. Nevertheless, the classification hyperplane is not located on the middle of two separating areas. Because the blue class has more sample size than the red class, the separating hyperplane is therefore pushed towards the red class. This will lead to the performance degeneration when confronted with new data. In this paper, we propose a novel linear binary classification that works well on HDLSS without considering any dimensionality reduction, denoted by population-guided large margin classifier (PGLMC). Figure 1(e) demonstrates that our method not only addresses the data-piling issue, but also preserves the distribution pattern of the original data as much as possible. Our method is stable; no matter the samples are balanced or not across two classes.

In the HDLSS data set, we should observe that the borders of the samples for two classes are unstable, even sometimes stochastic, as reported in [8, 26-29]. In Fig 2, we use a two-dimension example to illustrate this variability. All data from two classes are generated from a multivariate normal distribution $N_d(\pm\mu,\Sigma)$, where the dimension $d = 2$, $\mu = (1,2.5)$ and $\Sigma = [1\ 0;0\ 1]$. The Bayes rule in this example has the direction $w_{Bayes} = \Sigma^{-1}(2\mu)$, and the Bayes intercept is $b_{Bayes} = 0$. In Fig. 2(a), there are 50 samples in positive (blue) class and 10 samples in negative (red) class, $m = 5$. The border-based classic method, SVM, presents its decision boundary as the cyan line, which is deviated heavily from the Bayes rule direction. The mean vector of positive samples is denoted as the blue solid dot, and the mean vector of negative samples as red solid dot. When we increase the negative samples to 20, $m = 2.5$, SVM gives a visible change of decision boundary as shown in Figure 2(b). However, despite the increase of the samples, the mean vectors for two classes vary very little. In this sense, the first-order statistic (mean vector) of the samples for two classes is more robust than border points for learning or classification task. Inspired by this observation, our motivation is to combines the local structure of the hyperplane and the global statistics information of training samples to construct a new object function. In Fig. 2, it is visible that our proposed PGLMC preserves a good and stable boundary to distinguish the samples in two classes which is not subject to the imbalanced sample sizes.

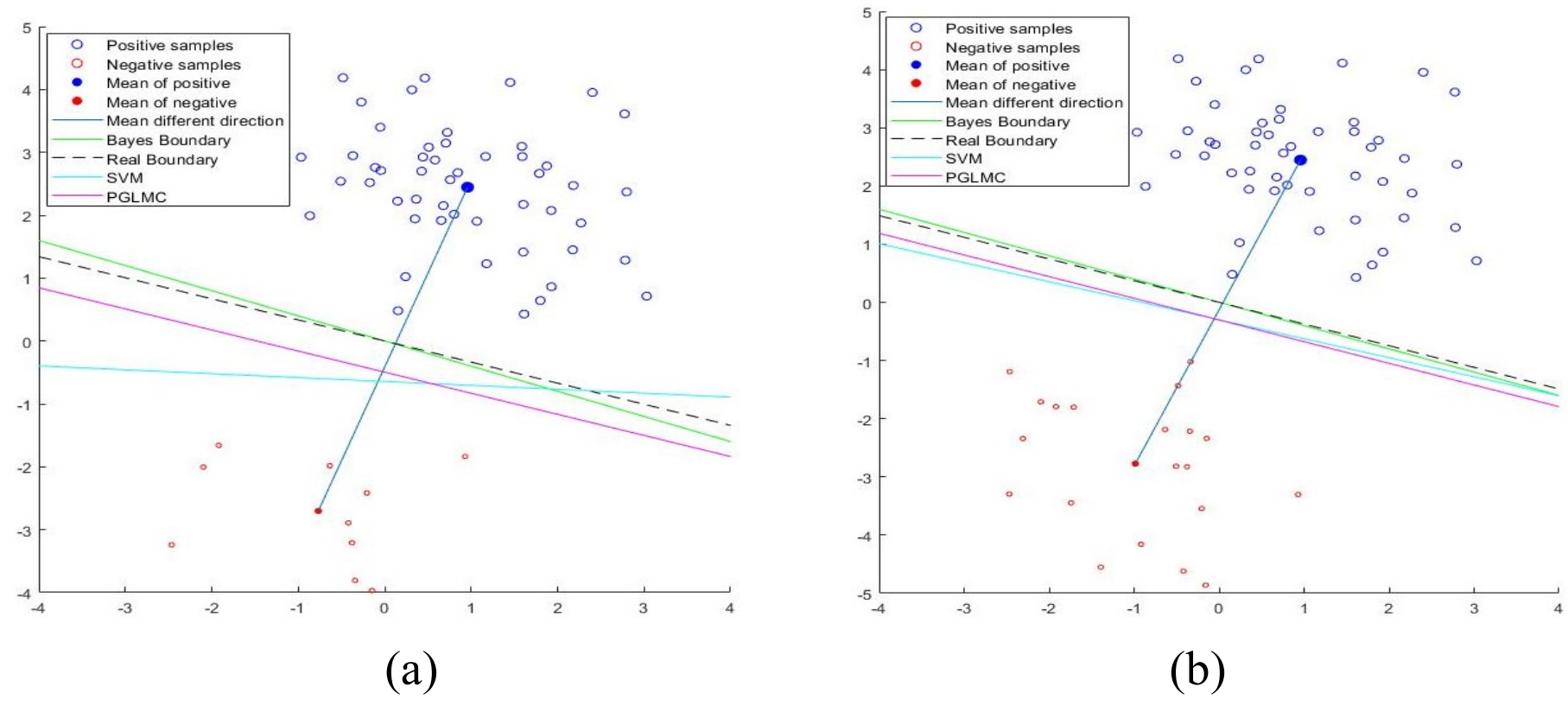

Figure 2. The illustration of border variability for two classes in HDLSS. (a) 50 samples in positive class and 10 samples in negative class. (b) 50 samples in positive class and 20 samples in negative class.

In this paper, we design a novel linear binary classifier, population-guided large margin classifier (PGLMC), which works well on HDLSS without considering the dimensionality reduction. In addition to dealing with HDLSS and imbalanced data, the dual formulation of PGLMC is relatively easy to implement and holds much lower computational complexity than the Distance-weighted based methods owing to solve the similar Convex Quadratic Programming formulation as in SVM. We prove in theory that PGLMC is with three characteristics, (1) Fisher consistency, (2) Asymptotic under extremely imbalanced data, (3) Asymptotic with infinite dimensions. PGLMC is robust to the model specification, which makes it very popular in various real application, especially for HDLSS. Furthermore, linear classifiers are the most important and the most commonly used classifiers, as they are often easy to interpret in addition to reasonable classification performance [4]. The linear classifiers can be easily extended to nonlinear forms and the kernel space. Even though our main focus is on HDLSS data, PGLMC can be applied to general data with arbitrary dimensions (or the imbalanced data) in diverse applications (e.g., DNA classification, digital recognition, medical image analysis, face recognition, and so on), as shown with several experiments in Section 4.

The rest of this paper is organized as follows. Section 2 briefly introduces the related methods in HDLSS, especially SVM and the methods based on Distance Weighting. Section 3 elaborates the proposed PGLMC. Section 4 demonstrates experimental results. Finally, Section 5 concludes the paper.

## 2. Related Work

The relative study on HDLSS problem are reviewed, which are categorized to two groups as mentioned in the above. In the first group, there are two strategies for dimensionality reduction. One is to project original high dimensional feature space to a low dimensional space using transform technologies prior to classification. Typical such transform technologies are principal component analysis (PCA), LDA, and manifold learning. These methods require some proper prior assumptions when applied. For example, LDA require the covariance within classes to be known apriori in order to establish a discriminant rule. However, in HDLSS, the sample covariance is ill-conditioned and hence induce singularity in the estimation of the inverse covariance matrix. By the specific restrictive assumptions on sample covariance, some improved methods by regularity were proposed to address the singularity of sample covariance or select valuable features in HDLSS, i.e., Null Space Linear Discriminant Analysis (NLDA) [12], Modified LDA(MLDA) [30], Diagonal Linear Discriminant Analysis (DLDA) [31], Diagonal Quadratic Discriminant Analysis (DQDA) [31], Nearest Shrunken Centroids (NSC) [32], Shrinkage-based DQDA (SDQDA) [33] and Regularized max-min linear discriminant analysis(RMMLDA) [15]. Another strategy in the first group to dimensionality reduction is to integrate feature selection or inducing sparsity in common classifiers [34]. A large number of feature selection methods and regularization for sparsity were proposed, such as filter methods, wrapper methods, embedded methods (Ridge Regression [35], LASSO [36], SCAD [37], Elastic-net [34]) and so on. Therefore, the classifiers in this type on HDLSS depends mostly on the performance of dimensionality reduction. Even when sample size is very small and dimensionality of features is large, such as $n < 8$ and $d \gg n$, most of the above mentioned methods cannot work properly.

In the second group, SVM and the methods based on Distance Weighting work well without dimensionality reduction on HDLSS data set. Here, we will introduce briefly them, which are the widely-used methods and have received a lot of attention. These methods have been extended to kernel forms. Here, we are only concerned with linear implementations in the whole paper.

For binary classification problems, the classification rule is to map a data point in sample space to a class label chosen from $Y$, $\phi{:}X \to Y$, where $X \in R^d$ and $Y \in \{+1, -1\}$. A linear discriminant function can be denoted as $f(x) = \left(w^T x + b\right)$, where the coefficient direction vector $w \in R^d$ has unit $L_2$ norm, and $b \in R$ is the intercept term. The classification rule is then $\phi(x) = sign(f(x))$, that is, the sample space $R^d$ is divided into halves by the separating hyperplane defined by $\left\{x{:}f(x) = \left(w^T x + b\right) = 0\right\}$.

**2.1 SVM**

The SVM is defined to the minimization problem as below:

$$\underset{w,b,\varepsilon_i}{\operatorname{argmin}} \frac{1}{2}\|w\|^2 + C\sum_{i=1}^{N} \varepsilon_i \tag{1}$$

$$\text{s.t. } y_i\left(w^T x_i + b\right) \geq 1 - \varepsilon_i,\ \ \varepsilon_i \geq 0, i = 1,2,\cdots,N$$

SVM is a classic method and has a good performance record as a powerful classification tool. It possesses some nice properties [6]: (1) Its dual formulation is relatively easy to implement through Quadratic Programming (QP), (2) SVM is robust to the model specification, which makes it very popular in various real applications. (3) it is not sensitive to imbalance data. But, it may suffer from a loss of generalizability in HDLSS settings due to the data-piling phenomenon[20]. Data-piling indicates a type of overfitting.

**2.2 The Methods based on Distance Weighting**

(1) *Distance-weighted discrimination*

Distance-weighted discrimination (DWD) is a recently developed classifier which aim to improve the SVM in the HDLSS setting[6, 19]. It also maximizes a notion of gap between classes: the DWD method maximizes the harmonic mean of the distances of all data vectors to the separating hyperplane.

The objective function of DWD:

$$\underset{w,b}{\operatorname{argmin}} \sum_{i=1}^{N} \left(\frac{1}{r_i} + C\eta_i\right) \tag{2}$$

$$\text{subject to } r_i = y_i\left(x_i^T w + b\right) + \eta_i,\ r_i \geq 0, \eta_i \geq 0, \quad \|w\|^2 \leq 1$$

While DWD overcomes the data-piling phenomenon or overfitting to a certain extent, it has caused new issues: (1) The intercept term $b$ of the DWD method is sensitive to the sample size ratio $m$ between the two classes, that is, to the imbalanced data[20]. (2) Computing consumption. The current state of the art implementation for DWD is based on second-order cone programming (SOCP), which is more computationally demanding than quadratic programming[23].

(2) *Weighted Distance-weighted discrimination*

To get over the imbalanced data, the weighted DWD (wDWD) [20] was proposed to incorporate class proportions as well as prior costs to improve upon standard DWD

$$\underset{w,b}{\operatorname{argmin}} \sum_{i=1}^{N} \mathcal{W}(y_i)\frac{1}{r_i} \tag{3}$$

$$\text{subject to } \|w\|^2 \leq 1, r_i \geq 0,\ r_i = y_i\left(x_i^T w + b\right), i = 1,2,\cdots,N.$$

Where $\mathcal{W}(y_i)$ is the weight for the *i*th training data point. The solution to (8) is totally determined by the ratio of $\mathcal{W}(+1)$ and $\mathcal{W}(-1)$, instead of the exact values of the two weights. Although weighted DWD has improved standard DWD for imbalanced data and various nonstandard situations, wDWD is still with heavily computing consumption due to SOCP.

## 3. Population-Guided Large Margin Classification Machine

Following the discussion from the previous section, we know that (1) SVM only maximize the gap between 2 classes to find the supporting vectors and the projecting direction $w$ in the solution space. It implies that SVM merely exploit the local structural information of the training samples without considering the statistics of training samples. When the sample size is large, the training samples can reflect the actual distribution or overall structure of the whole population, and the local structural information expressed by the supporting vectors is sufficient to construct an appropriate classification projecting direction $w$ and intercept term $b$. However, when the sample size is small, local structural information expressed by the supporting vectors is biased and not sufficient to obtain the fitting projecting direction $w$ and intercept term $b$. (2) DWD and wDWD, on the other hand, maximize the harmonic mean of the distances of all data vectors without considering the local structural information.

Based on the previous discussion and other results in the literature [6, 8, 26], here, a linear classifier is proposed (conceived) with a projecting direction $w$ given by the comprehensive consideration of local structural information of the hyperplane (or gap) and the statistics of training samples (or population). The reasons why we apply the sample mean as representative statistic of population are explained in Supplementary Material. This proposed method is called as the population-guided large margin classifier (PGLMC). The objective function is as follow:

$$\underset{w}{\operatorname{argmin}}\left(\frac{\|w\|^2}{(m_1-m_2)^T w}+C_0\sum_{i=1}^{N}\xi_i\right) \tag{4}$$

$$\text{s.t. } y_i\left(w^T x_i+b\right)\geq 1-\xi_i,\ i=1,2,\cdots,N$$

$$\xi_i=H_1(y_i f(x_i)),\ (m_1-m_2)^T w\geq 2$$

Where $m_i$ is the mean of training samples from $i$th class, $i = 1,2$. Term $\|w\|^2$ is to maximize the gap between two classes. At the same time, the term $(m_1 - m_2)^T w$ is to control the centroids of training samples from two classes as far as possible along the projecting direction $w$. The hinge loss $H_1(u) = \ell(u) = (1-u)_+$, where $(u)_+ = u$ if $u \geq 0$ and 0 others. When $(m_1 - m_2)^T w = 2$, the equation (4) will be equivalent to the expression of SVM, in other words, SVM is a special case of formula (4).

The Lagrangian formulation is given by

$$L(w,b,\lambda,\alpha) = \frac{\|w\|^2}{(m_1-m_2)^T w} + C_0 \xi + \lambda[2 - (m_1 - m_2)^T w] + \alpha^T[\mathbf{1} - Y(Xw + b\mathbf{1})] - \boldsymbol{\mu}^T \cdot \xi \qquad (5)$$

where $Y$ is the $n \times n$ diagonal matrix with the components of $y_i$ on its diagonal; $X \in R^{n\times p}$, which ith row is sample $x_i$; $w \in R^{p\times 1}$ is the direction vector or projecting vector; $\mathbf{1}$ is the column vector 1; $\alpha = (\alpha_1,\cdots,\alpha_N)^T \in R^{N\times 1}$, $\alpha_i > 0$ and $\alpha_i$s are Lagrangian multipliers; $\xi = (\xi_1,\cdots,\xi_N)^T \in R^{N\times 1}$ ; $\lambda \geq 0$ is a Lagrangian multiplier for second term.

The motivation of formulation (4) has two folds, (1) it is desired that $\|w\|^2$ is as small as possible, which ensures that the gap (minimum distance) between two classes is as large as possible, (2) the item of denominator $(m_1 - m_2)^T w$ will be as large as possible to force the projecting direction $w$ to be the one, on which the distances between the projecting points from two classes are as far as possible. From this assumption, it can be found that in the above formulation, $\frac{1}{(m_1-m_2)^T w}$ and $\left[2 - (m_1 - m_2)^T w\right]$ play a similar role. Therefore, these two items would be turned into be a new item $\left[C - (m_1 - m_2)^T w\right]$ (the detailed proof is provided in the supplemental files). The optimization objective function can be reformulated so as to facilitate calculation as below:

$$L(w,b,\lambda,\alpha) = \frac{1}{2}\|w\|^2 + C_0 \xi + \lambda[C - (m_1 - m_2)^T w] + \alpha^T[\mathbf{1} - \xi - Y(Xw + b\mathbf{1})] - \boldsymbol{\mu}^T \cdot \xi \; (6)$$

By differentiating the Lagrangian formulation with respect to $w$, $b$ and $\xi$, we can get the following conditions:

$$\frac{\partial L}{\partial w} = w - \lambda(m_1 - m_2) - X^T Y^T \alpha = 0 \tag{7}$$

$$\frac{\partial L}{\partial \xi} = C_0 - \boldsymbol{\alpha} - \boldsymbol{\mu},\ C_0 = \boldsymbol{\alpha} + \boldsymbol{\mu} \tag{8}$$

$$\frac{\partial L}{\partial b} = \alpha^T Y \mathbf{1} = 0 \tag{9}$$

When substituting (7), (8) and (9) into (6), we can obtain the dual form as follow

$$L(\lambda,\alpha) =-\frac{1}{2}\left\{[\lambda,\alpha^T]\begin{bmatrix}(m_1 - m_2)^T(m_1 - m_2) & (m_1 - m_2)^T X^T Y^T \\ YX(m_1 - m_2) & YXX^T Y^T\end{bmatrix}\begin{bmatrix}\lambda \\ \alpha\end{bmatrix}\right\} + [\lambda,\alpha^T]\begin{bmatrix}C \\ \mathbf{1}\end{bmatrix} \tag{10}$$

When provided that

$$\beta = \left[\lambda,\alpha^T\right]^T, A = \begin{bmatrix}(m_1 - m_2)^T(m_1 - m_2) & (m_1 - m_2)^T X^T Y^T \\ YX(m_1 - m_2) & YXX^T Y^T\end{bmatrix}$$

$A$ is a symmetric positive semidefinite matrix, $\tau = [C\ \mathbf{1}]^T$. The equation (6) can be

$$L(\beta) =-\frac{1}{2}\left\{\beta^T A \beta\right\} + \beta^T \tau \tag{11}$$

So, the optimization problem (4) can be transformed into the following

$$\underset{\beta}{\operatorname{argmax}}\, L(\beta) \tag{12}$$

$$\text{s.t. } \beta^T \begin{bmatrix}0 & 0 \\ 0 & Y\end{bmatrix} \mathbf{1} = 0, C_0 \geq \alpha \geq 0, \lambda \geq 0$$

It can be known that the Karush-Kuhn-Tucker condition must be satisfied as follow:

$$C_0 \geq \alpha_i \geq 0$$

$$y_i\left(w^T x_i + b\right) - 1 \geq 0$$

$$\alpha_i\left[y_i\left(w^T x_i + b\right) - 1\right] = 0$$

Therefore, derived from the above formula, the concept of support vector is still applicable. The support vector set is defined as same as the one in support vector machine, $S = \left\{x_i \middle| y_i\left(w^T x_i + b\right) = 1\ and\ \alpha_i > 0\right\}$.

The intercept term $b$ can be obtained as follows:

$$b = \frac{1}{|S|} \sum_{x_i \in S} \left( y_i - w^T x_i \right) \tag{13}$$

Given a new sample $x$, the classifier of PGLMC is defined by

$$\hat{y} = f(x) = \left( w^T x + b \right) = \left[ \lambda(m_1 - m_2) + X^T Y^T \alpha \right]^T x + b \tag{14}$$

The performance and asymptotic properties of PGLMC can be explored in three different flavors: (1) the $d$ fixed and $n \to \infty$ (Fisher consistency), (2) the $d$ and $n_+$ fixed, $n_- \to \infty$ (Asymptotic under extremely imbalanced data), (3) $n$ fixed, $d \to \infty$. Detailed information about the theoretical properties and computation complexity of PGLMC is discussed in the Supplementary material.

# 4. Results

To evaluate the proposed PGLMC algorithm, in this section, we perform a series of experiments systematically on both simulation data and real-world classification problems. First, we present 2 synthetic datasets for clearly comparing PGLMC with DWD, wDWD and SVM. Second, on real-world problems, 6 datasets depicted in Ref [2-4, 38, 39] are used to evaluate the classification accuracies derived from PGLMC in comparison to the other algorithms in the HDLSS framework.

In the following experiments, to eliminate the dependence of the results on the particular training data used, some measures are defined and the average (or mean) of these measures are reported, which are obtained for different randomly sample splits. And the programs were developed in MATLAB and R, and executed in AMD Athlon(tm) X4 730 Quad Core Processor 2.8G Hz system with 16GB RAM. For the methods based on Distance weighting, we adopt the linear binary implementations in R package 'kerndwd'[23].

## 4.1 Measures of Performance

In order to evaluate the performance of PGLMC and compare it fairly with other methods, I adopted the performance measures, some of which were used in reference[6]. In general, the Bayer rule classifier was as the benchmark in HDLSS for comparison. In the below simulation setting, we suppose that data are sampled from two multivariate normal distribution with different mean vectors ($m_+$ and $m_-$) and same covariance matrices $\Sigma$. We can get the following Bayes rule

$$sign\left(x^T w_{Bayes} + b_{Bayes}\right) \tag{15}$$

$$w_{Bayes} = \Sigma^{-1}(m_+ - m_-)$$

$$b_{Bayes} =- \frac{1}{2} w^T (m_+ + m_-)$$

Then, four performance measures are defined as follows:

(1) The *correct classification rate* (CCR): a measure of classification performance when the balanced data for two classes is involved

$$CCR = \frac{the\ number\ of\ correct\ classified\ samples}{the\ total\ of\ samples\ for\ testing} \tag{16}$$

Remark: when imbalanced data sets are involved, this measure is biased, i.e., it is possible that while CCR is very high, almost all of the samples belonging to minority class are misclassified.

(2) The *mean within-class error* (MWE) for an out-of-sample test set

$$MWE = \frac{1}{2n_+} \sum_{i=1}^{n_+} \delta\left(y_i^+ = \hat{y}_i\right) + \frac{1}{2n_-} \sum_{j=1}^{n_-} \delta\left(y_j^- = \hat{y}_j\right) \tag{17}$$

Where $y_i^+$ denotes the sample $x_i$ belongs to class $Y =+ 1$; $n_+$ and $n_-$ are the sample size in positive and negative class; $\delta(\cdot)$ is the delta function which is equal to 1 when $\delta(0)$, and 0 when otherwise. The measure can work well while the imbalanced data are involved.

Remark: when $n_+ = n_-$, $MWE = CCR$.

(3) The *deviation* of the estimated intercept $b$ from the Bayes rule intercept $b_{Bayes}$: $|b - b_{Bayes}|$

(4) *Angle* between the estimated discrimination direction $w$ and the Bayes rule direction $w_{Bayes}$: $\angle(w, w_{Bayes})$

**4.2 Simulations Data**

We conduct 2 experiments using simulated data to compare the classification and interpretability performance among the PGLMC, the original SVM, DWD and wDWD. The classification performance is measured for a large test data set with 4000 observations (2000 for each class). The interpretability is a vague concept, and we measure it by the angle between the discriminant direction vectors for classifier under investigation and for the Bayes classifier. In general, it is right that the closer to the Bayes rule direction, the better the interpretability is. The process is in accordance with the literature [6, 21].

Let us consider two different simulation settings: (1) independent, (2) block-interchangeable covariance. In each setting, samples from two classes are generated from multivariate normal distributions $N_d(\pm \mu, \Sigma)$.

(1) Independent structure: constant mean difference, identity covariance matrix example. $\mu \equiv c\mathbf{1}_d$, and $\Sigma \equiv I_d$, where $c > 0$ is a scaling factor which makes $2c\|1_d\|_2 = 2.7$. This corresponds to the Mahalanobis distance between the two classes and represents a reasonable difficulty of classification using the Bayes rule.

(2) Block-interchangeable structure: decreasing mean difference, block-diagonal interchangeable covariance matrix. Let $\mu \equiv c\nu_d$ with $\nu_d = (\sqrt{50}, \sqrt{49}, \cdots, \sqrt{1}, 0, \cdots, 0)^T \in R^d$, and $\Sigma \equiv \mathrm{Block}_{\mathrm{Diag}}\{\Sigma, \Sigma, \cdots, \Sigma\}$, where each $\Sigma$ is a $50 \times 50$ interchangeable sub-covariance matrix whose diagonal entries are all 1 and off-diagonal entries are 0.8. The scaling factor $c$ is chosen to make the Mahalanobis distance $\left\{(2cv_d)\Sigma^{-1}(2cv_d)^T\right\}^{1/2} = 2.7$ . Here, $\text{Block}_{\text{Diag}}\{\Sigma_1,\Sigma_2,\cdots,\Sigma_k\}$ means that a matrix is constructed whose diagonal elements consist of $\Sigma_1$, $\Sigma_2$, $\cdots$, $\Sigma_k$ in sequence.

In both simulation settings, we let the positive class sample size be 200 and the negative class sample size be 50, (i.e., the training data are imbalanced with imbalance factor $m = 4$). We vary the dimension $d$ in {100, 200, 300, 500, 1000}, thus last three cases correspond to HDLSS setting.

4.2.1 Experiment 1

In Figure 3, we report the comparison results of 10 replications for independent structure among DWD, wDWD, SVM and PGLMC applied to a test dataset with 2000 samples in each class, which are generated according to the constant mean difference, identity covariance matrix example. The boxplots in Figure 3(a) show the scatter intervals of the correct classification rate for 4 methods. The boxplots of DWD express the larger ranger of CCR and worst classification performance than those of other methods. Even when the dimension is 200, the CCR of DWD is only 50%. wDWD gets the best CCR, and PGLMC follows it, both of which are better than SVM. However, while the dimensionality increases, the performance of CCR for wDWD, SVM and PGLMC gradually tend to be consistent. Figure 3(b) is about the mean within-group errors for 4 methods. It is obvious that according to the measure of MWE, the wDWD is the best; PGLMC is the second; SVM is better than DWD. Figure 3(c) is about the angle between the estimated discrimination direction w and the Bayes rule direction for 4 methods. In the case of low dimensions, PGLMC is optimal; while the dimensionality increases, these methods are gradually becoming consistent. Figure 3(d) is about the deviation of the estimated intercept b from the Bayes rule intercept. The deviation of PGLMC has a larger spreading interval than those of others when dimensions are smaller than 300. Moreover, while the dimensionality increases, the intercept of PGLMC converge very soon. But, provided that all of these measures in Figure 3 are analyzed simultaneously, it can be found that (1) angle and deviation are not strict indicators of classification performance; (2) the classification performance of PGLMC is not sensitive to the intercept; (3) The CCR, WME and angle of PGLMC gradually converge to those of SVM as the dimension increases.

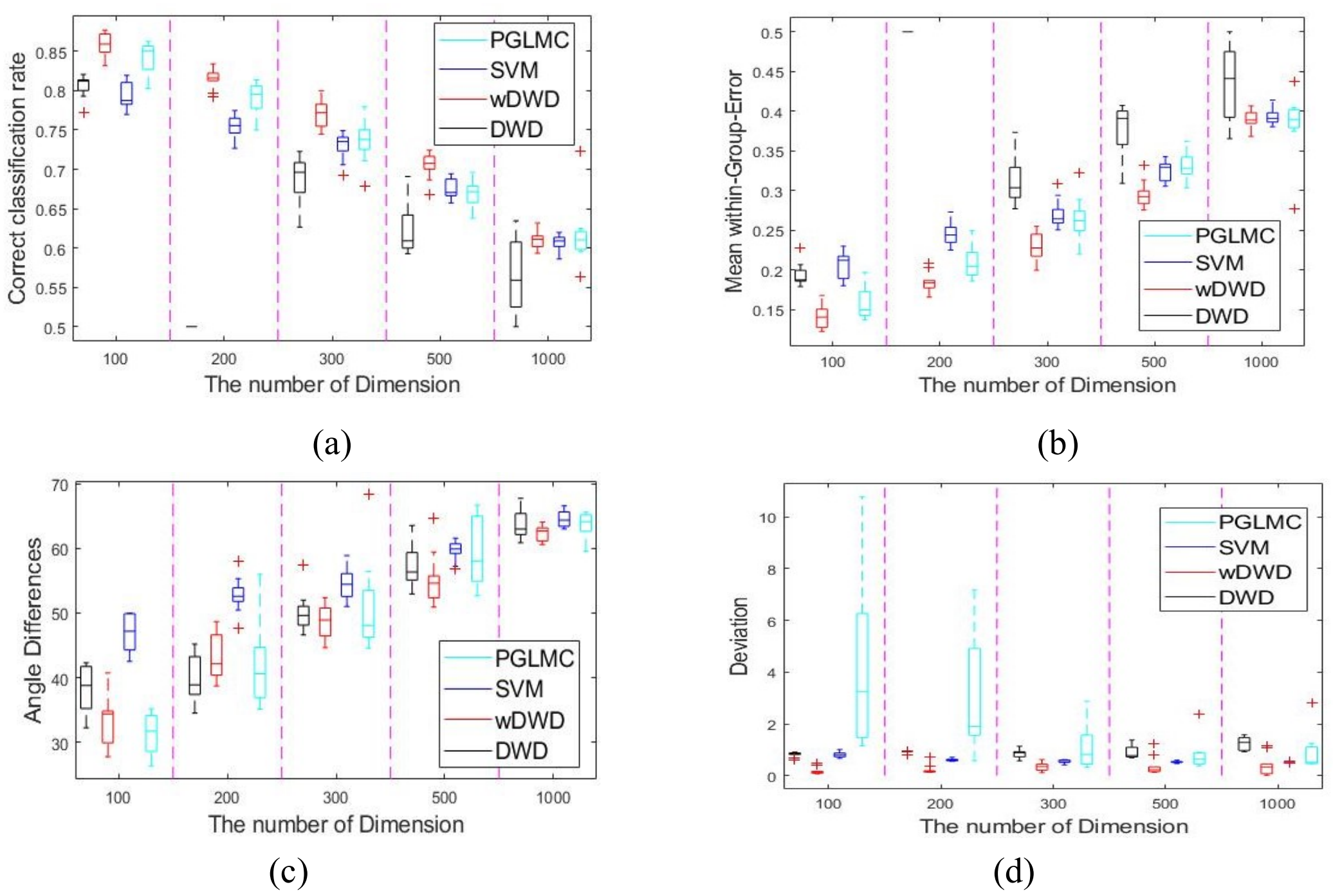

Figure 3. Comparison among four methods for simulation experiment 1 with 10 replications. (a) the boxplots of CCRs. (b) The boxplot for MWE. (c) The boxplot for the angle differences. (d) The boxplot for the deviation.

### 4.2.2 Experiment 2

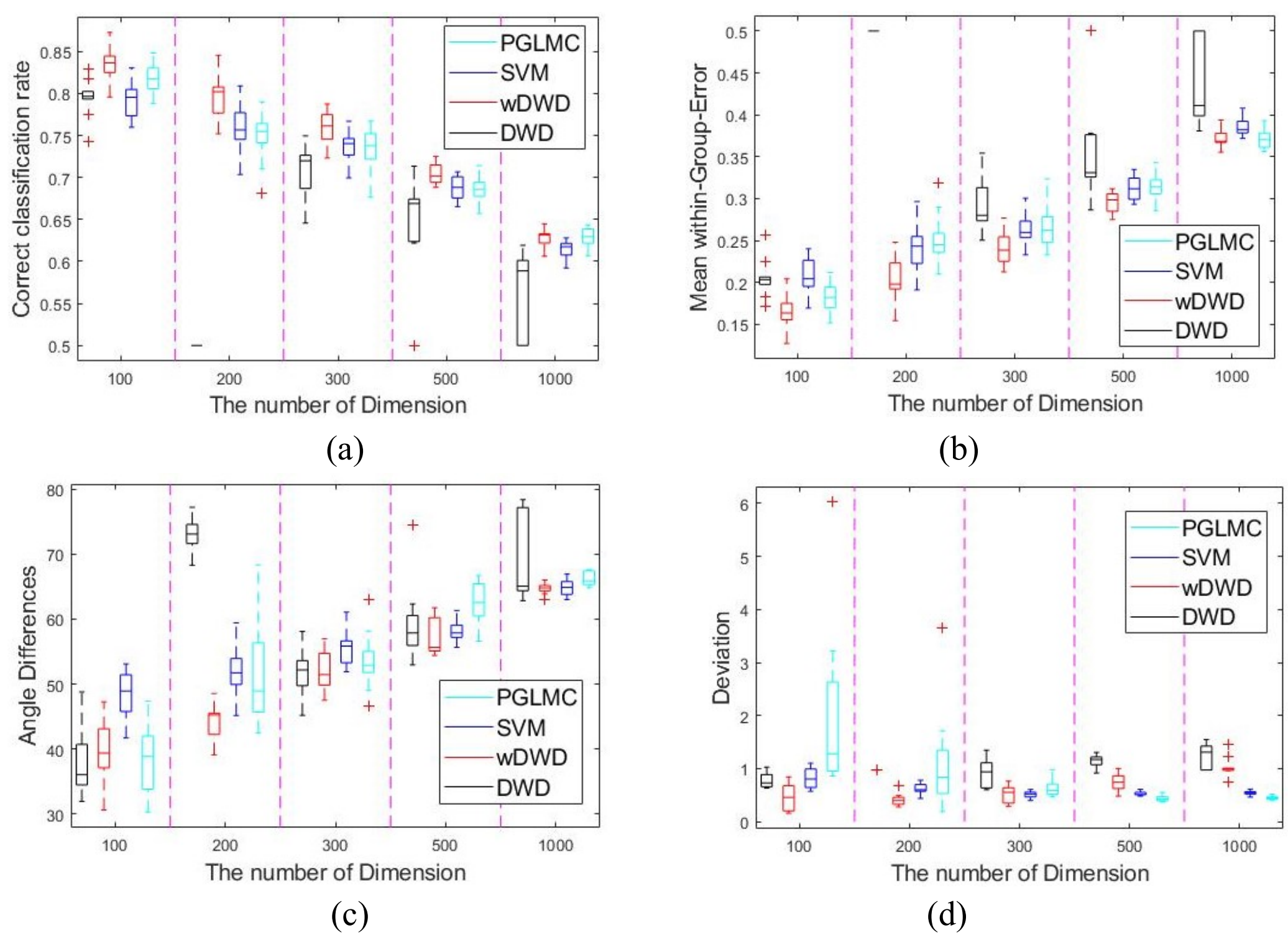

Figure 4. Comparison among four methods for simulation experiment 2 with 100 replications. (a) The boxplots of CCRs (b) The boxplot for MWE. (c) The boxplot for the angle differences. (d) The boxplot for the deviation.

We have conducted the same comparison for the decreasing mean difference, block-diagonal interchangeable covariance matrix example in experiment 2. In Figure 4, we display the comparison results with 100 replications for simulation experiment 2 among DWD, wDWD, SVM and PGLMC applied to a test dataset with 2000 samples in each class, which are generated according to the same distribution with training samples. In this test, in whole, the classification performance of wDWD overperforms others. PGLMC is the second, and its correct classification rate or MWE is slightly better than SVM. DWD is the worst one. For DWD, even when the dimension is 200, its classification direction vector is seriously deviated from Bayes direction so that only 50% for correct classification rate is obtained. From Figure 4(a-b), for dimension 1000, PGLMC is the best one. For the deviation of the intercepts, the deviations of DWD and wDWD are approximately monotonic increasing with the increase of dimension. However, on the contrary, the intercept deviations of SVM and PGLMC are monotonic decreasing. While the dimension is larger than 500, the intercepts deviations of SVM and PGLMC are stable, and PGLMC's intercept is closer to the Bayes baseline.

On simulation datasets, the performance of PGLMC is second only to the wDWD, and suboptimal and superior to other methods (DWD and SVM).

**4.3 Real applications**

In this subsection, we demonstrate the performance of PGLMC which are compared with the competing classifiers on 6 real data sets: (1) UCI Wine [38, 40]. (2) The Human Lung Carcinomas Microarray dataset, which has been analyzed earlier in [2]. (3) The Golub dataset [3]. (4) The Optical Pen Digitals data set [38, 40]. (5) The Kidney radiomics dataset [4]. (6) Face image data : EYaleB [39, 40].

According to the characteristics of the datasets, we split the experiments into three subsections: classic low dimensional dataset, high dimensional datasets and image data sets. Table 1 shows the characteristics of the data sets used in the experiments. These real-world applications rang from general physical properties of objects, gene expression data classification to image based disease discrimination, handwritten digits recognition and face recognition. We consider that (1) the data dimensions range from 13 to 3051, (2) the sample sizes range from 23 to 2414, (3) the number of classes range from 2 to 38. Here, when the number of samples in the data sets are much fewer than the dimensionality of each samples (i.e. n << d), the data set is definitely with high dimension. Because our focus is only binary linear classification, when there are more classes than 2 in the data set, we iterate to select one class as positive and the rest classes as negative in the experiments. Thus, we report the mean

CCR and MWE as the results. For each data set, we report detailed data preparation, evaluation methods and comparison results.

Table 1. Characteristics of the Data Sets Used in the Experiments

| Type | Data Set | Dim | Class | Examples | Comments |
|---|---|---|---|---|---|
| Low Dim | Wine | 13 | 3 | 178 | Multi-class, small sample size |
| High Dim | Lung | 200 | 4 | 56 | Multi-class, small sample size |
| | Golub | 3051 | 2 | 38 | Small sample size |
| Image data | Optical pen | 64 | 10 | 177 | Multi-class, medium sample size |
| | kidney radiomics | 182 | 2 | 23 | Small sample size |
| | EYaleB | 1024 | 38 | 2414 | Medium sample size |

4.3.1 Experiment 3: Wine data set

The Wine data set is from UCI machine learning repository [38]. To test the accuracy of classification, we adopt 5-fold cross-validation to evaluate the correct classification rate (CCR) and mean within-group error (MWE) for the two classes over 100 random splits. In each split, there are 3 steps to select the parameter and evaluate the performance: (1) 4-fold samples were for training, and one-fold samples as testing data. (2) The samples within training data were still divided into 4-folds, 3-folds to obtain the models, the rest one to test in training. (3) Select an optimal from 4 models to conduct the evaluation on the testing data. For each split, we can get 5 results for CCR and MWE when 5-fold cross-validation are adopted. The means of these 5 results are as CCR and MWE for this split. The seed for splitting is fixed to maintain the same splits for all methods.

In this data set, there are 3 classes. Thus, we iterate to select one class as positive and the rest classes as negative in each split, and evaluate the performances for each class. The sample size is 59, 71 and 48 with imbalance factor $m \in \{2.01, 1.51, 2.71\}$, respectively.

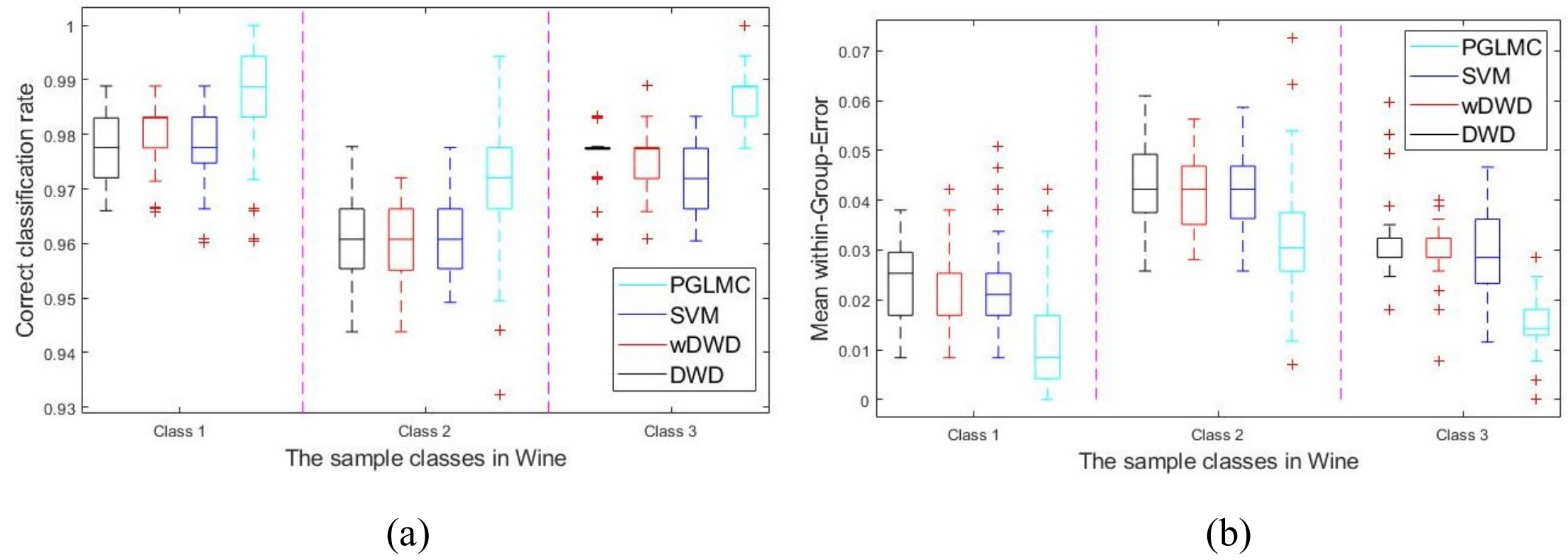

(a) (b)

Fig. 5. Comparison among four methods on Wine data set with 100 replications. (a) The boxplots of CCRs. (b) The boxplot for MWE.

Table 2. The performance on Wine data set

| Methods | CCR | MWE |
|---|---|---|
| DWD | 97.22 | 3.25 |
| wDWD | 97.24 | 3.01 |
| SVM | 97.11 | 3.12 |
| PGLMC | 98.27 | 1.90 |

Figure 5 shows the comparison results with 100 replications on Wine data set. As shown in Fig. 5, we find that (1) PGLMC achieves the best mean performance whether CCR or MWE, (2) DWD, wDWD and SVM get the similar results in most cases. The mean value of CCR and MWE over 100 replications are summarized in Table 2.

4.3.2 Experiment 4: Lung data set

The Human Lung Carcinomas Microarray Data set contain 4 classes: pulmonary carcinoid, colon, normal and small cell carcinoma, with sample sizes of 20, 13, 17 and 6, respectively. We combine the subclasses, pulmonary carcinoid and normal as the positive class; then the subclasses, colon and small cell carcinoma, to form negative class. The sample size is 37 and 19 with imbalance factor $m \approx 1.95$. The original data contain 12,625 genes measured using the Affymetrix 95av2 GeneChip. We first filter genes using the ratio of the sample standard deviation and sample mean of each gene and keep 200 of them with large ratios [2, 31, 41].

To test the accuracy of classification, we adopt 5-fold cross-validation to evaluate the correct classification rate and within-group error for the two classes over 100 random splits. In each split, there are 3 steps to select the parameter and evaluate the performance: (1) 4-fold samples were for training, and one-fold samples as testing data. (2) The samples within training data were still divided into 5-folds, 4-folds to obtain the models, the remaining one for validation. (3) Select the optimal model from the 5 models to conduct the evaluation on the testing data. For each split, we can get 5 results for CCR and MWE when 5-fold cross-validation are adopted. Because the true Bayes rule is unknown, we cannot get the measures of deviation and angle.

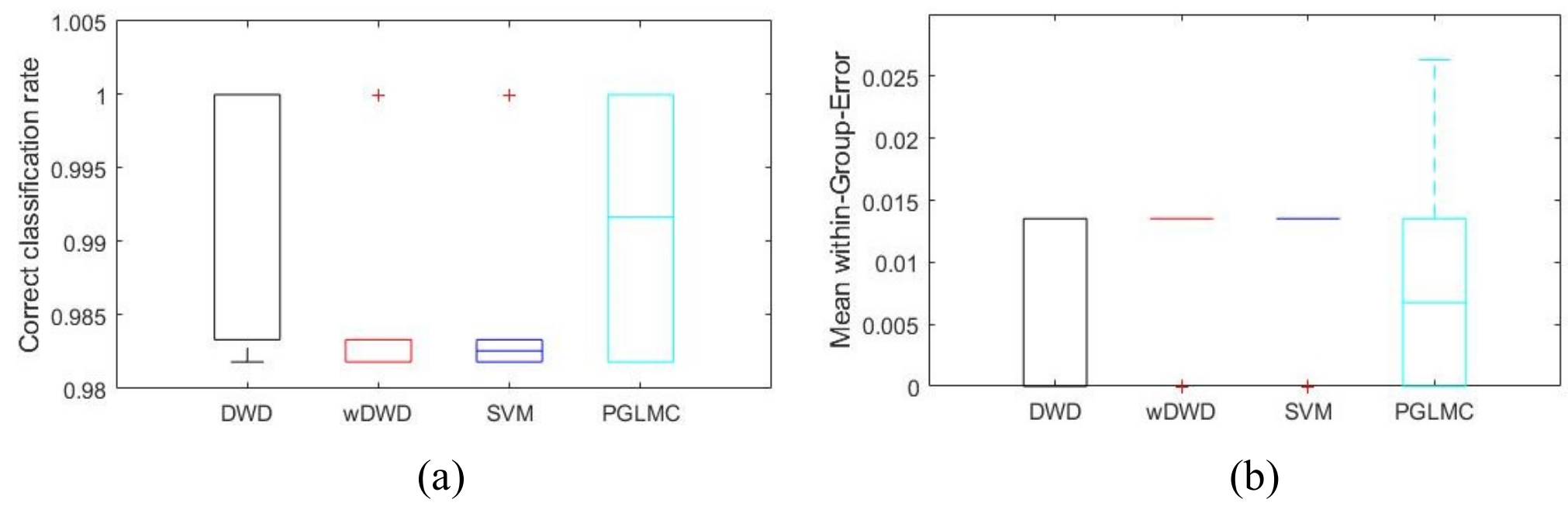

Figure 6. Comparison among four methods on Lung dataset with 100 replications. (a) The boxplots of CCRs. (b) The boxplot for the mean within-class error.

Figure 6 shows the comparison results with 100 replications for 4 methods on the Lung dataset in experiment 3. Because m≈1.95, the results derived from CCR are a little different from that of WME. However, it is clear that (1) DWD and PGLMC overperform wDWD and SVM, (2) DWD and PGLMC have similar performance while DWD has a slight superiority over PGLMC due to the whisker in Fig6(b).

4.3.3 Experiment 5: Golub data set

In this experiment, the Golub data set is used, which contains 38 tumor mRNA samples and 3051 genes from the leukemia microarray study [3]. The pre-processing was done as described in [31].

The sample size is 27 and 11 for 2 classes with imbalance factor $m \approx 2.45$. Therefore, we expect the PGLMC and SVM will give better result than DWD because the latter is subject to the imbalanced sample size. As done in the experiment 4, we split all specimens to five folds, in which 4-folds are for training and one fold for testing. Parameters for each method are tuned via 5-fold cross-validation within training data. This process is repeated for 100 times. The CCRs and MWEs are exhibited in Fig. 7.

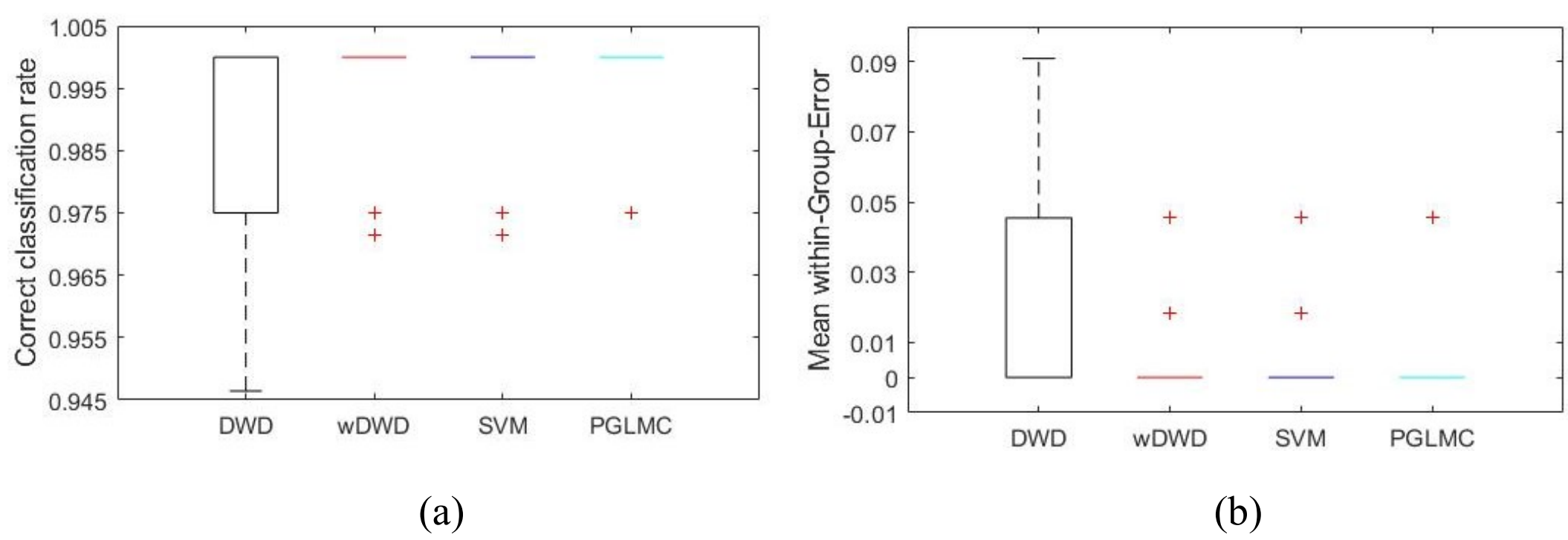

Figure 7. Comparison between four methods on Golub data for 4 methods from example 4 with 100 replications. (a) The boxplots of CCRs. (b) The boxplot for the mean within-class error.

As shown in Figure 7, in this data set, PGLMC is the best one than other methods (DWD, wDWD and SVM), especially superior to DWD. PGLMC have almost perfect performances.

### 4.3.4 Experiment 6: Optical pen digitals

The data set, Optical pen digitals, from UCI machine learning repository [38], composes of the normalized 32x32 bitmaps of handwritten digitals (from 0 to 9) from a preprinted form. 1797 digital images were involved in this recognition processing. The digital examples are shown Supplementary Material Fig S1. For features extraction, 32x32 bitmaps are divided into nonoverlapping blocks of 4x4 and the number of on pixels are counted in each block yielding 64 features for each digital image. About the detailed pre-processing steps, please refer to [38].

As was done in [40], we split all specimens to five folds, four of which are for training and one fold for testing. Parameters for each method are tuned via 5-fold inner cross-validation within training data. This process is repeated for 10 times. The CCRs and MWEs are exhibited in Fig. 8.

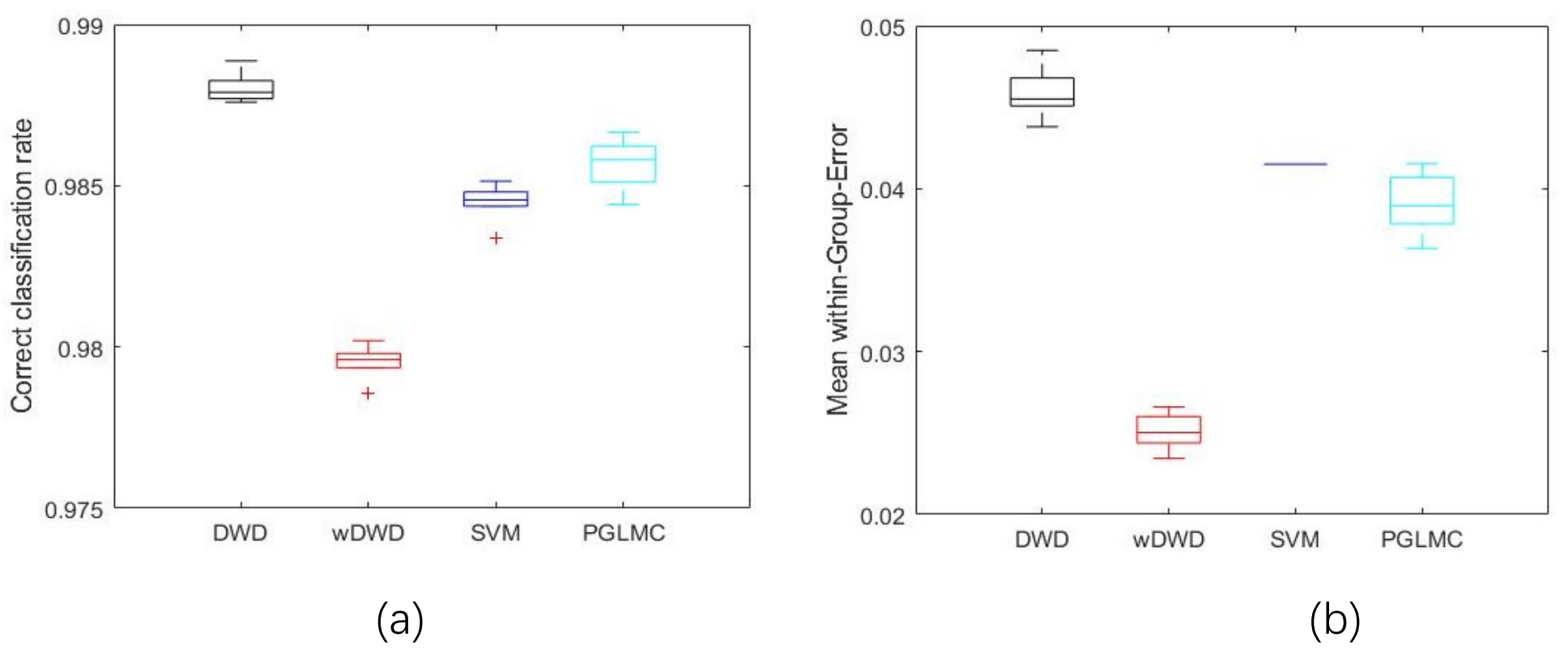

Figure 8. Comparison among four methods on Optical Pen digital dataset with 10 replications. (a) The boxplots of CCRs. (b) The boxplot for the mean within-class error.

Figure 8 illustrates the comparison results with 10 replications for 4 methods on the Optical Pen digitals dataset. In Fig.8(a), DWD gets the best CCR, PGLMC is second, and wDWD is the fourth. But, in Fig.8(b), we observe amusing change. wDWD obtains the lowest MWE, and PGLMC is still second, and DWD get the worst MWE among these four methods. It is reasonable when imbalanced factor $m = 9$ for each digit is considered. PGLMC displays its superiority on both of CCR and MWE.

4.3.5 Experiment 7: Kidney radiomics data set

The kidney radiomics data set is extracted from MRI images and PET images of the patients with renal cancer which have 23 radiomics samples with 182 dimension observations (feature vector), and the respective molecular subtype (ccA or ccB) of clear cell renal cell carcinoma (ccRCC) as the predictor. In this data set, original images per subject involved four sets of high-resolution 3D images for radiomics analysis, including multiphasic dynamic contrast enhancement (DCE), Dixon fat images (Dixon_F), Dixon water images (Dixion_W), and 2-[fluorine-18]fluoro-2-deoxy-D-glucose (18F-FDG) PET, as shown in Supplementary Material Fig S2. The sample size is 14 and 9 from both classes with imbalance factor $m \approx 1.56$. For more details and pre-processing (i.e. multimodality registration, marked tumor sampling and feature extraction) about this data set, please refer to [4].

Because of few of samples in this dataset, we use leave-one-out for performance evaluation. The training process is same with that of experiment 4 and experiment 5. 5-fold cross-validation within training data is used to select the model parameters. The whole process is repeated for 10 times.

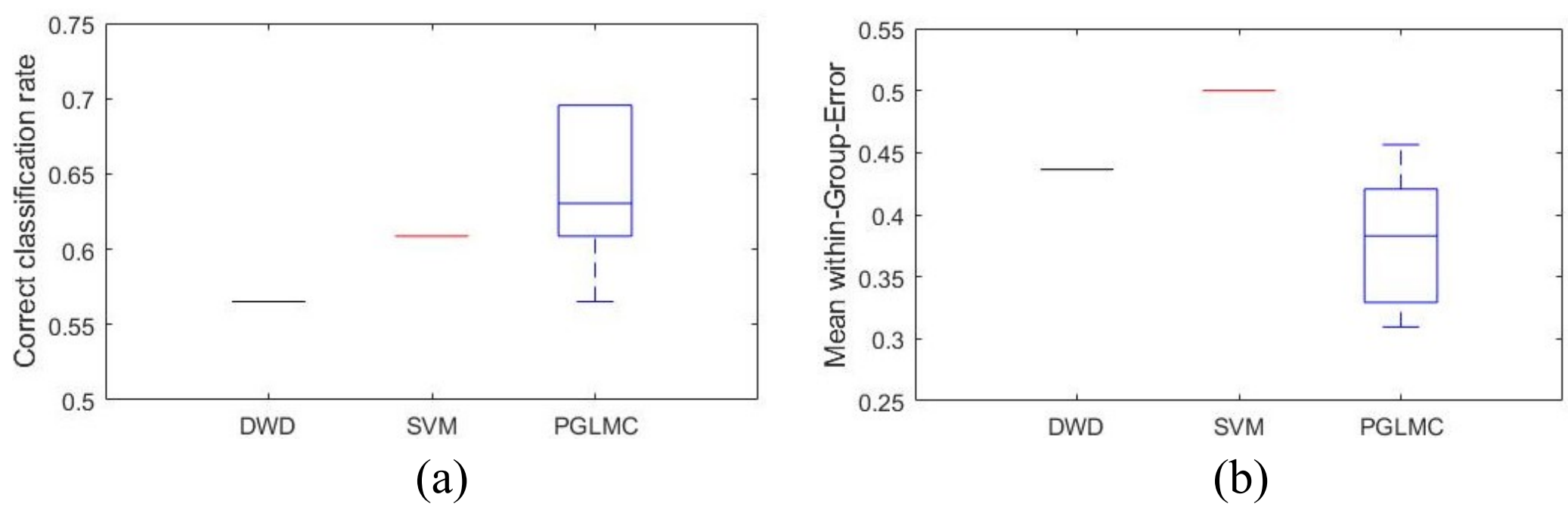

Figure 9. Comparison among four methods on kidney radiomics dataset with 10 replications. (a) The boxplots of CCRs. (b) The boxplot for the mean within-class error.

In Fig. 9, we give the comparison results of box plots for DWD, SVM and PGLMC because the codes of wDWD do not work well on this small size dataset. On this specimen dataset, PGLMC is with overwhelming superiority over other methods whether in CCR or MWE. Although the CCR of SVM is about 60% and better than that of DWD in Fig. 9(a), SVM actually misclassifies all samples of negative class (9 samples). Therefore, MWE displays more reasonable results in Fig. 9 (b) that is DWD better than SVM in this small and imbalanced dataset.

4.3.6 Experiment 8: EYaleB data set for Face Recognition

The task of face recognition always involves many classes and a large of image features (high dimension). In the EYaleB data set, there are 2414 images for 38 individuals. Each individual contains around 64 near frontal images under different illumination conditions, and have diverse expressions, as exhibited in Supplementary Material Fig S3 for a few examples. All images are cropped and resized to $32 \times 32$ pixels as done in [39, 40]. The features used in this experiment are the original gray values in the images. Therefore, the dimension of feature vectors for a sample of each class is 1024.

To compare the performance in the follow-up subsection with the state-of-the-art methods on this data set in [40], we adopt 10-fold cross-validation on the whole data set over 10 random splits. The seed is fixed to maintain the same split for each method. Because Cheng has reported the detailed results for SVM on this data set [40], we just conduct the tests for DWD, wDWD and PGLMC. As shown in Fig.10, PGLMC obtains the best CCR over DWD and wDWD. The wDWD obtains the lowest MWE, and PGLMC is still suboptimal while imbalanced factor $m = 37$ for each individual. An extensive comparison on this data set will be reported in the next subsection.

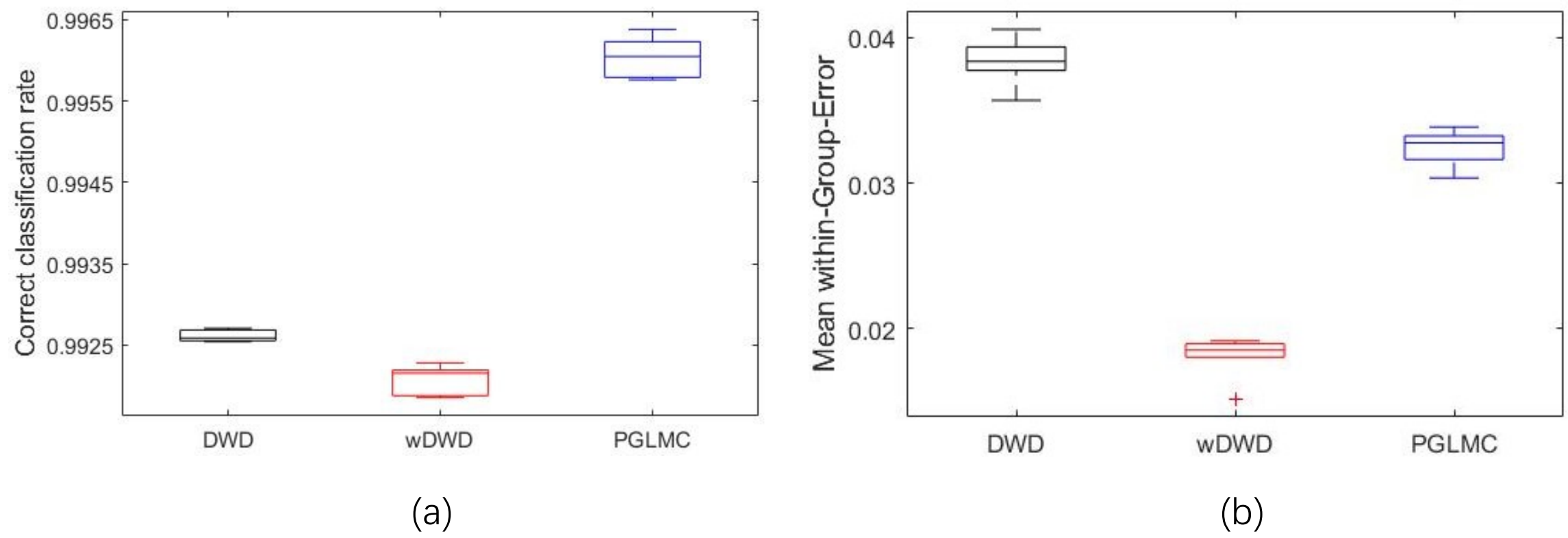

Figure 10. The boxplot for comparison among 3 methods on EYaleB face data set. (a) The boxplots of CCRs. (b) The boxplot of MWE.

4.3.7 Further analysis of real datasets

Here, since the many state-of-the-art methods were tested on different data sets, we group the above data sets to 2 types, (1) data sets generally for HDLSS methods, i.e. Lung data set, Golub data, Optical pen digitals set and Kidney radiomics data set, (2) data sets which were extensively used to test the performance of general classifiers, including Wine data set and EYaleB face data set.

For type 1, we calculate the correct classification rate (CCR) for DWD, wDWD, SVM and the proposed PGLMC on four real datasets (Lung data set, Golub data set, Optical pen digitals and Kidney radiomics). In order to make a detailed analysis, we summarize the above experimental results in Table 3, in which the best results are marked in red and the suboptimal in cyan. It is shown that in experiment 4, DWD obtains the highest mean correct classification rate of 99.63% with 5-fold cross-validation and repetitions for 100 random splits; PGLMC obtains the suboptimal result 99.30% with the same condition as DWD; in experiment 5, the best mean CCR 99.90% is obtained by PGLMC, and SVM takes the second; in experiment 6, PGLMC gets the similar rank (suboptimal) of CCR and MWE with experiment 4; in experiment 7, PGLMC get the similar rank (optimal) of CCR and MWE with experiment 5, such that the best mean CCR 63.91 of PGLMC and the second 60.87 of SVM. Since these datasets are imbalanced, we can get another rank of MWE. However, it is clear that PGLMC can be the best one or approximate on each real HDLSS dataset. Especially, while there are only rare specimens in the dataset, PGLMC still work well. To express this more intuitively, we sum up the CCR and MWE of 3 experiments for real HDLSS datasets in Fig. 11.

Table 3. 4 real datasets and means of the experiment results for CCR and MWE

| Experiment | Data set | | number of | CCR of Methods (%) | | | | MWE (%) | | | |
|---|---|---|---|---|---|---|---|---|---|---|---|
| | samples | dimensions | random splits | DWD | wDWD | SVM | PGLMC | DWD | wDWD | SVM | PGLMC |
| Experiment 4 | Lung[2]: 37 positive; 19 negative | 200 | 100 | 99.63 | 98.43 | 98.39 | 99.30 | 0.35 | 1.25 | 1.23 | 0.73 |
| Experiment 5 | Golub[3]: 27 positive; 11 negative | 3051 | 100 | 98.83 | 99.40 | 99.63 | 99.90 | 1.99 | 0.43 | 0.56 | 0.18 |
| Experiment 6 | Optical: 1797 samples M=37 | 64 | 10 | 98.80 | 97.95 | 98.45 | 98.57 | 4.59 | 2.51 | 4.15 | 3.91 |
| Experiment 7 | Kidney[4]: 14 positive; 9 negative | 182 | 10 | 56.52 | — | 60.87 | 63.91 | 43.65 | — | 50 | 37.98 |

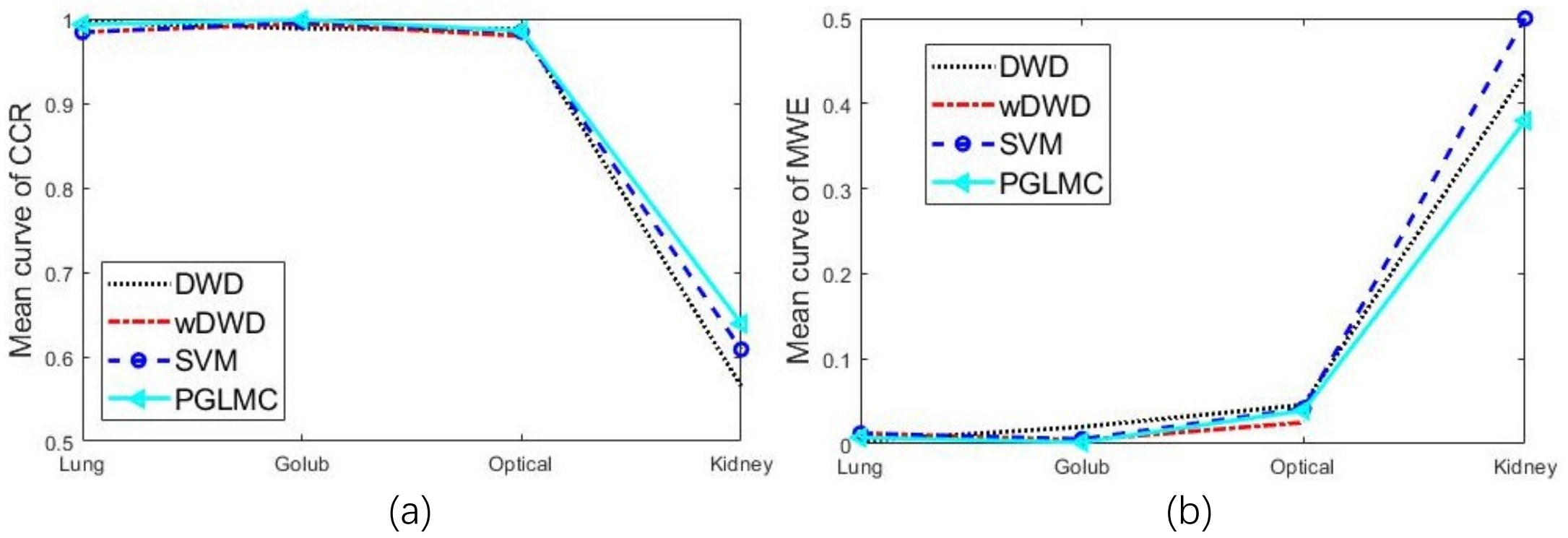

Figure 11. Comparison among four methods on 4 real datasets. (a) The mean curve of CCRs. (b) The mean curve for the mean within-class error.

For type 2, the Wine data set and EYaleB are often used to test the performance of general classifiers, involved the linear and nonlinear, binary or multiclass methods. Table 4 gives the comprehensive comparisons with the methods as reported in [40], which contained General Multiplicative Distortions (GMD) [40], Support Vector Machine(SVM) [17], Random Forest(RF) [42], KNN, Kernalized LDA(KLDA) [43], Kernel Minimax Classifier(MMC) [40] and SRC [44]. The GMD with radial basis function (RBF) kernel was denoted by GMD(R). SVM with polynomial was denoted as SVM (P), and with RBF kernels as SVM(R). KNNs with Euclidean and Cosine distance were denoted as KNN(E) and KNN(C). KLDA with polynomial and RBF kernels were denoted as KLDA(P) and KLDA(R), respectively. PGLMC achieves the best results on both of Wine and EYaleB data sets, although some nonlinear methods have been applied.

Table 4. Comparison of the correct classification rate (CCR) for PGLMC and other classifiers

| Data | GMD | GMD(R) | SRC | RF | SVM(P) | SVM( R) | KNN(E) | KNN(C) | KLDA(P) | KLDA(R) | MMC | DWD | wDWD | PGLMC |
|---|---|---|---|---|---|---|---|---|---|---|---|---|---|---|
| Wine | 97.28 | 95.41 | 88.21 | 97.28 | 97.76 | 98.24 | 71.57 | 83.73 | 97.28 | 97.17 | 92.22 | 97.40 | 97.40 | **98.27** |
| EYaleB | 95.61 | 98.96 | 93.64 | 99.10 | 93.51 | 94.59 | 97.87 | 98.04 | 93.87 | 96.87 | — | 99.74 | 99.73 | **99.76** |

From the above analysis, these results exhibit that although PGLMC is conceived for HDLSS, it is applicable to a variety of classic data sets, and overperforms the other classifiers (even nonlinear classifiers) in the most cases or get the comparable results despite PGLMC is only a linear binary classifier.

## 5. Conclusion

For the HDLSS problem, the existing methods (such as SVM, DWD and wDWD) are subject to certain disadvantages. In this paper, with analysis on the state of the art methods for HDLSS, we first introduce the idea, which guide the design of classification by combining the local structure of the hyperplane and the global statistics information of training samples (population). The objective function, Equation (4), optimize a combination of numerators and denominators. The numerator item $||w||^2$ ensures that the margin between two classes is as wide as possible. The denominator item $(m_1 - m_2)^T w$ uses the first order statistics of training samples to force the projecting direction $w$ to be the one, on which the distances between the projecting points from two classes are as far as possible. Based on the structure design, this proposed method possess the merits of both SVM and DWD.

The major advantages of this study were four-fold. First, it is not sensitive to the intercept term $b$. Second, it can work well with imbalanced data. Third, its dual formulation is relatively easy to implement. Its corresponding optimization problem follows the same Quadratic Programming formulation as SVM, rather than the SOCP in the DWD related methods. Fourth, it is robust to the model specification, which makes it very popular in various real application, especially for HDLSS.

We conduct study to explore the performance and asymptotic properties of PGLMC in three different flavors: (1) the d fixed and $n \to \infty$ (Fisher consistency), (2) the d and $n_+$ fixed, $n_- \to \infty$ (Asymptotic under extremely imbalanced data), (3) $n$ fixed, $d \to \infty$. These theory analyses accounts for the performance that we have seen in the experiments and assure that PGLMC can work well on the HDLSS and imbalanced datasets by combining the local structure of the hyperplane and the global statistics information.

In term of the above analysis and discuss, the PGLMC not only has the ideal characteristics in theory, but also hold much lower computational complexity than the DWD based methods owing to solve the similar Quadratic Programming formulation as SVM, rather than the Second-order cone programing in the DWD related methods. The experiment results manifested the superiority of PGLMC compared to the state-of-art algorithms (SVM and the DWD based methods) in HDLSS. Actually, these exhibit that it is a very promising linear binary classification, which is with great potential in many applications regardless of HDLSS. The PGLMC can be extended to multi-class classification, just as was reported for SVM and DWD[45-48]. Other extension studies are in progress, such as kernel PGLMC and sparse PGLMC.

## Acknowledgement

The authors would like to thank Prof. J.S. Marron in Department of biostatistics, University of North Carolina at Chapel Hill; Dr. Boxiang Wang in statistics at University of Minnesota; Prof. Xingye Qiao in Department of Mathematical Sciences, Binghamton University, for their kind help and discussion about the methods based on Distance weighting.

## Additional Information

**Competing financial interests:** The authors declare no competing financial interests.